\documentclass[]{inkmind}

\usepackage[toc,page,header]{appendix}
\usepackage{calc}
\usepackage{float}

\usepackage{minitoc}
\usepackage{graphicx}
\usepackage{subcaption}
\usepackage{pifont}
\usepackage{makecell}
\usepackage{booktabs}
\usepackage{multirow}
\usepackage{float}
\usepackage{enumitem}

\usepackage{mathrsfs}
\usepackage{adjustbox}
\usepackage{multirow}
\usepackage{multirow}
\usepackage{multicol}
\usepackage{tcolorbox}
\usepackage{changepage}
\usepackage{graphicx}
\usepackage{amsmath}
\usepackage{amssymb}
\usepackage{array}
\usepackage{bm}
\usepackage{tabularx}
\usepackage{colortbl}
\usepackage{hyperref}
\usepackage{afterpage}
\usepackage{minitoc}

\newcolumntype{C}{>{\centering\arraybackslash}X}

\newcommand{\fittowidth}[1]{%
  \sbox0{#1}%
  \ifdim\wd0>\textwidth
    \resizebox{\textwidth}{!}{\usebox0}%
  \else
    \usebox0%
  \fi
}

\title{Dyn-3D: Unveiling and Resolving Ego-Motion Ambiguity in Vision-Language Models}
\author{InkMind Team}
\abstract{
As Vision-Language Models (VLMs) tackle dynamic 3D spatial reasoning, ego-motion perception becomes essential to resolve monocular scale ambiguity. However, current models often overfit to smooth trajectory priors rather than genuinely understanding physical motion. Consequently, their spatial reasoning degrades severely under large displacements, a phenomenon we term Kinematic Collapse. This failure stems from spurious visual-motion correlations in natural videos and a lack of explicit physical supervision. To evaluate this, we introduce Dyn-3D, a benchmark using counterfactual 3D rendering to rigorously decouple visual changes from true kinematic properties. Furthermore, we propose the TempoVista framework, featuring the Kinematic-GSPO algorithm. By embedding metric physical ground truth into policy optimization, TempoVista explicitly grounds visual representations in 3D space. Experiments demonstrate that our approach significantly improves both motion estimation and robust spatial reasoning by utilizing camera dynamics as an effective geometric calibration signal.
}

\date{September 1, 2026}
\checkdata[Project Page]{\url{https://inkmind-ai.github.io/Dyn-3D/}}
\checkdata[GitHub]{\url{https://github.com/InkMind-AI/Dyn-3D}}

\begin{document}

\maketitle

%不需要目录就注释掉 注意目录不要和第一页放在一块 要有\newpage
\newpage
\tableofcontents
\newpage

\section{Introduction}

\afterpage{%
\begin{figure*}[t]
	\centering
	\includegraphics[width=0.96\textwidth]{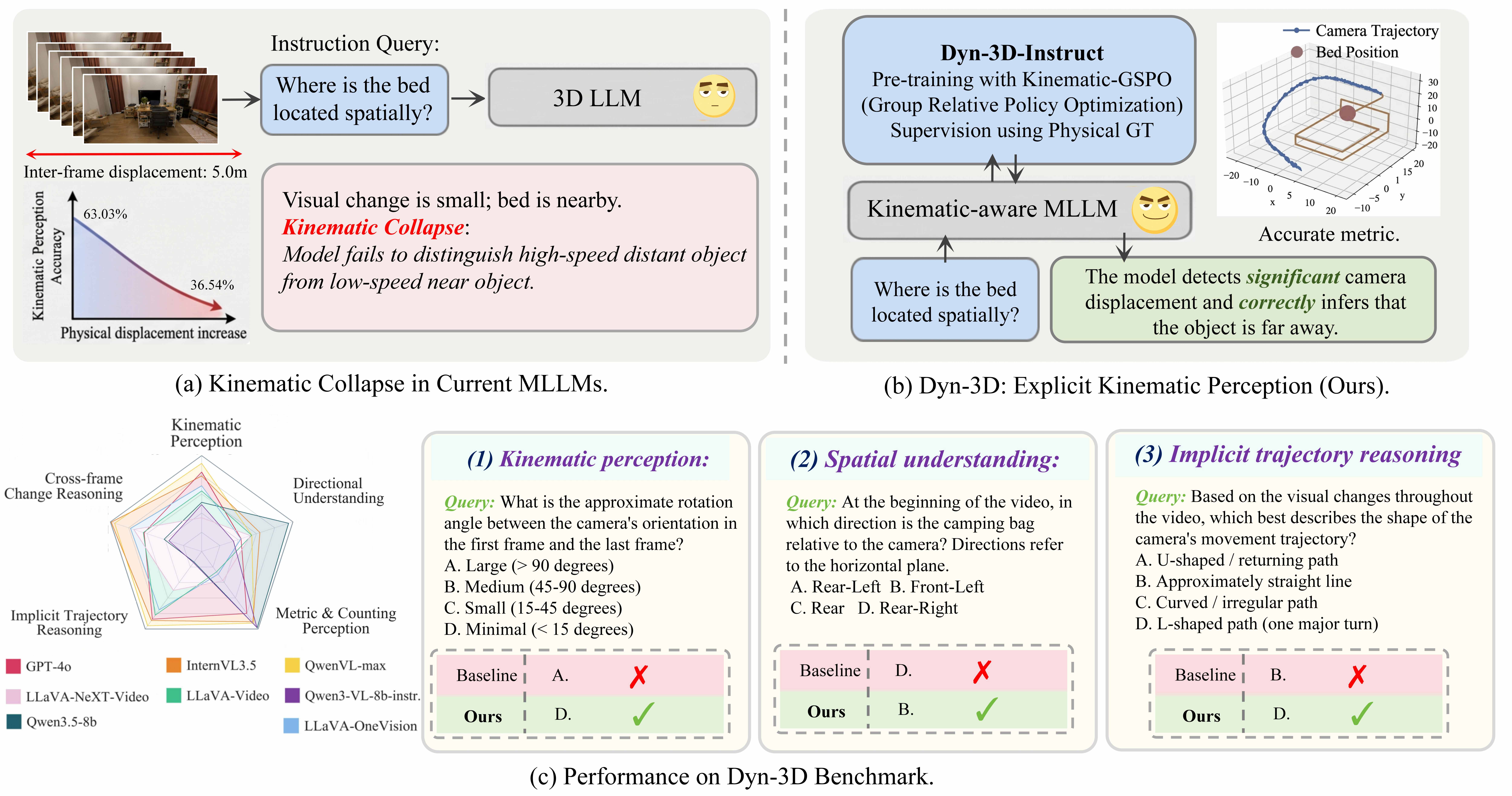}
	\caption{\textbf{Kinematic Collapse and the Dyn-3D Benchmark.} \textbf{(a) Kinematic Collapse:} Current MLLMs rely on superficial visual changes rather than genuine physical displacement. \textbf{(b) Dyn-3D:} Supervised by physical ground truth via Kinematic-GSPO, our model achieves explicit kinematic perception for accurate spatial and trajectory reasoning. \textbf{(c) Performance:} Our approach significantly outperforms standard baselines across diverse spatiotemporal tasks on the Dyn-3D benchmark.}
	\label{fig:teaser}
\end{figure*}%
}

Multimodal Large Language Models (MLLMs)~\cite{zhang2024vision} are transitioning from static 2D image perception to dynamic 3D spatial understanding. In unstructured physical environments, intelligent agents must recognize object semantics and parse 3D spatial logic. This geometric cognition, which maps 2D pixels to metric depth in the 3D physical world, is essential for embodied intelligence~\cite{gupta_embodied_2021} and complex physical interactions.

To achieve accurate video spatial understanding, models must first comprehend camera motion. In continuous video streams, pixel variations on the 2D image plane result from the coupling of 3D scene geometry and viewpoint changes. Without knowing the camera trajectory and motion magnitude, models cannot determine whether parallax and scaling are caused by the actual spatial depth of objects or by fast camera movement. Because monocular visual signals have inherent scale ambiguity, they cannot directly provide absolute depth. Therefore, ego-motion perception serves as a metric reference connecting the 2D visual stream with 3D space. Models can only obtain reliable geometric constraints by accurately parsing how parallax changes with their own displacement.

However, current models struggle to understand camera motion. Although some models demonstrate strong spatiotemporal capabilities on standard video benchmarks~\cite{li2024mvbench,fu2025video}, this largely stems from overfitting to smooth trajectory priors in training data. The dense sampling rates and slow motion characteristics of mainstream datasets cause models to learn spurious correlations between visual changes and physical displacements. Consequently, their spatial cognition fails when encountering large displacements or non-smooth trajectories, a phenomenon we term Kinematic Collapse.
For instance, in our pilot evaluation of Qwen3.5-9B~\cite{qwen2026qwen35}, increasing the inter-frame physical displacement from 0.1 to 5.0 meters drops overall accuracy by 25.30 percentage points (from 63.97\% to 38.67\%). More importantly, the capability to perceive camera motion plummets under high dynamics: kinematic perception accuracy falls from 69.71\% to 9.66\%, and implicit trajectory reasoning remains below the 25\% chance level across all displacement magnitudes. This failure stems from kinematic ambiguity. Without explicitly modeling camera extrinsics or velocity, models easily confuse visual similarity with spatial proximity, thereby losing the metric anchor required for accurate 3D spatial decoding.

To systematically evaluate the ability of Vision-Language Models (VLMs) to understand real camera motion, we introduce the Dyn-3D Benchmark. To our knowledge, this is the first benchmark designed to assess ego-motion perception of VLMs within a 3D spatial understanding framework. In natural videos, visual and motion streams are highly coupled, as large motions usually accompany large optical flows, making models prone to relying on these spurious correlations. To address this, the Dyn-3D Benchmark utilizes the controllable re-rendering capabilities of 3D Gaussian Splatting~\cite{kerbl_3d_2023} to synthesize counterfactual video pairs with identical visual paths but different kinematic characteristics. This dataset contains 16,063 test samples. Empirical results show that even advanced models exhibit near-chance accuracy when estimating camera motion magnitude under non-smooth or counter-intuitive trajectories, confirming a blind spot in kinematic perception.

To address these issues, we introduce the Dyn-3D-Instruct dataset (33.6K samples) and propose the Kinematic Group Sequence Policy Optimization (Kinematic-GSPO) algorithm. Since standard next-token prediction lacks explicit physical constraints, Kinematic-GSPO applies process-level supervision using physical ground truth. By penalizing unphysical motion inferences, it compels the model to decouple true kinematic features from visual appearances. Consequently, the model accurately perceives complex camera trajectories and physical displacements. Experiments demonstrate that this internalized physical prior directly enhances robust global spatial understanding, yielding strong performance in inferring depth, orientation, and occlusion across multiple 3D spatial benchmarks.

Our main contributions are as follows:
\begin{itemize}
	\item We reveal model vulnerability in spatial understanding without smooth trajectory priors, attributing this failure to kinematic ambiguity.
    \item We propose Dyn-3D and Dyn-3D-Instruct via counterfactual re-rendering to train and evaluate VLM ego-motion perception.
	\item We leverage physical ground-truth supervision to explicitly enhance the capability of foundation models in comprehending camera motion.
	\item Empirically, improved kinematic perception significantly boosts model accuracy on 3D spatial benchmarks like Dyn-3D and VSI-Bench.
\end{itemize}

\section{Related Work}
\label{sec:appendix_related_work}

\paragraph{Video and spatial reasoning in MLLMs.}
Recent MLLMs have extended vision-language modeling from static images to videos and long-context inputs, achieving strong performance on broad video question-answering benchmarks such as MVBench, EgoSchema, and Video-MME~\cite{zhang2024vision,li2024mvbench,DBLP:conf/nips/MangalamAM23,fu2025video}. However, these benchmarks primarily assess event recognition, temporal localization, and commonsense reasoning, with limited emphasis on metric spatial inference. Spatially oriented benchmarks, including ScanQA~\cite{azuma2022scanqa}, SpatialVLM~\cite{chen2024spatialvlm}, OpenEQA~\cite{majumdar2024openeqa}, and VSI-Bench~\cite{yang2025thinking}, move closer to embodied 3D understanding by evaluating object-grounded question answering, metric distance estimation, and visual-spatial memory. More recent studies have further expanded this line of research to language-guided object grounding in 3D environments~\cite{ReferSplat,ding2026extrinsplat,ding2026zerosplat} and temporally object tracking in dynamic 4D scenes~\cite{li2026lmm}. Collectively, these efforts establish spatial reasoning as a central bottleneck for MLLMs. Nevertheless, they typically evaluate spatial understanding through downstream task performance without explicitly isolating ego-motion as an independent factor~\cite{liu2026egotl}. In natural videos, camera displacement is tightly entangled with parallax, optical flow, and changes in object appearance. Consequently, high benchmark accuracy may still arise from exploiting dataset-level motion correlations rather than explicitly recovering and reasoning about the underlying camera dynamics.

\paragraph{Ego-motion and geometric ambiguity.}
Classical multiple-view geometry treats camera motion and scene structure as coupled variables, and monocular video is scale-ambiguous without additional metric cues or priors~\cite{hartley2004multiple}. Structure-from-motion and SLAM systems address this ambiguity through explicit pose estimation, geometric verification, bundle adjustment, and map consistency~\cite{schonberger2016structure,mur2017orbslam2}. In contrast, VLMs are usually trained with visual-textual objectives and rarely receive direct supervision over camera extrinsics, velocity, or metric displacement. This creates a blind spot: large appearance changes can be mistaken for large translation, while stable appearance can hide substantial physical displacement. Dyn-3D targets this visual-kinematic ambiguity directly.

\paragraph{Controllable 3D rendering for diagnostic evaluation.}
Neural scene representations, from NeRF to 3D Gaussian Splatting, enable photorealistic novel-view synthesis from posed images~\cite{mildenhall_nerf_2022,kerbl_3d_2023}. Their controllability makes them suitable for counterfactual evaluation: one can preserve scene content and the spatial path while changing trajectory dynamics, or decouple rotation from translation under the same reconstructed geometry. Built on high-fidelity ScanNet++ scenes and nerfstudio-based 3DGS rendering~\cite{yeshwanthliu2023scannetpp,tancik2023nerfstudio}, Dyn-3D converts ego-motion from an uncontrolled nuisance factor into an explicit intervention variable. This design prevents models from solving spatial questions through superficial visual-motion correlations.

\paragraph{Reasoning optimization for multimodal models.}
Chain-of-thought prompting and recent RL-based reasoning methods improve model behavior by optimizing intermediate reasoning traces in addition to final answers~\cite{NEURIPS2022_9d560961,deepseekai2025deepseekr1incentivizingreasoningcapability}. Video-R1 extends this direction to video MLLMs~\cite{feng2025videor1reinforcingvideoreasoning}. Our setting requires a stricter form of process supervision: a reasoning trace is not sufficient because it is fluent; it must also be physically consistent with the underlying camera trajectory. Kinematic-GSPO therefore augments answer-level reward with explicit motion-consistency supervision over path length, displacement, direction, rotation, and speed. This makes geometric correctness an optimization target rather than an emergent byproduct.

\section{Pilot Experiments}
We design two pilot experiments to investigate whether models genuinely understand physical motion or merely rely on superficial visual changes.

\subsection{Kinematic Collapse Under High Dynamics}

We evaluate Qwen3.5-9B~\cite{qwen2026qwen35} by increasing the inter-frame physical displacement $\Delta d$ from 0.1m to 5.0m. The evaluation contains 16,063 samples, covering spatial understanding, kinematic perception, and implicit trajectory reasoning. Table~\ref{tab:pilot_delta} reports both the sample composition and the corresponding accuracy under different displacement magnitudes.
As $\Delta d$ increases, the overall accuracy drops by 25.30 percentage points
(from 63.97\% to 38.67\%), and spatial understanding accuracy falls from
67.53\% to 52.76\%. Crucially, kinematic perception accuracy plunges from
69.71\% to 9.66\%, and implicit trajectory reasoning remains below the 25\%
chance level across all displacement magnitudes.
This reveals that Vision-Language Models (VLMs) rely heavily on continuous
visual cues common in smooth training data. When large displacements break
local visual correspondences, their spatial reasoning fails. We define this
systematic degradation caused by increased motion magnitude as Kinematic
Collapse.

\begin{table}[t]
\centering
\small
\caption{Pilot evaluation of Qwen3.5-9B under increasing inter-frame physical displacement $\Delta d$. ``\#'' denotes the number of samples in each bucket; the five buckets partition the 16,063-question Dyn-3D benchmark. Accuracy is reported in \%. Chance level is 25.0\% for all four-option questions.}
\label{tab:pilot_delta}
\begin{tabularx}{\columnwidth}{l *{5}{C}}
\toprule
 & \multicolumn{5}{c}{\textbf{Inter-frame displacement} $\Delta d$} \\
\cmidrule(lr){2-6}
\textbf{Subset} & \textbf{0.1\,m} & \textbf{0.5\,m} & \textbf{1.0\,m}
                & \textbf{2.0\,m} & \textbf{5.0\,m} \\
\midrule
All (\#)         & 2969  & 1463  & 3845  & 5037  & 2749  \\
All Acc.         & 63.97 & 51.59 & 48.58 & 42.56 & 38.67 \\
\midrule
Spatial (\#)     & 2349  & 1051  & 2768  & 3576  & 1882  \\
Spatial Acc.     & 67.53 & 62.90 & 57.08 & 51.84 & 52.76 \\
\midrule
Kinematic (\#)   & 449   & 292   & 729   & 971   & 582   \\
Kinematic Acc.   & 69.71 & 30.71 & 31.29 & 23.04 & 9.66  \\
\midrule
Trajectory (\#)  & 171   & 120   & 348   & 490   & 285   \\
Trajectory Acc.  & 0.00  & 3.33  & 17.24 & 13.47 & 4.91  \\
\bottomrule
\end{tabularx}
\end{table}

\subsection{Visual-Kinematic Misalignment}

To avoid option-position bias, we construct four option-shuffled
evaluations for each of the 835 B17 motion-type recognition video-question pairs, resulting in 3,340 evaluation instances with balanced answer positions. These instances are derived from the B17 subset of the 16,063-question Dyn-3D benchmark rather than treated as additional benchmark questions.

% Results show that all models struggle on these decoupled trajectories. Qwen3.5-9B~\cite{qwen2026qwen35} achieves only 28.74\% accuracy on pure rotation and 5.69\% on pure translation, yielding a decoupled average of 17.22\%. Qwen3-VL-8B-Instruct~\cite{bai2025qwen3vltechnicalreport} and Qwen-VL-Max~\cite{alibabacloud2026qwenmodels} obtain similarly low decoupled averages of 13.70\% and 19.84\%, respectively. The error distribution reveals a consistent bias: models frequently classify pure rotation as translation plus rotation, with misclassification rates of 71.26\%, 99.70\%, and 100.00\% for the three models, respectively. They also confuse pure translation with either translation plus rotation or pure rotation. These results show that current VLMs conflate viewpoint change with physical displacement, confirming a severe Visual-Kinematic Misalignment.
Results show all models struggle on these decoupled trajectories. Qwen3.5-9B~\cite{qwen2026qwen35} achieves 28.74\% accuracy on pure rotation and 5.69\% on pure translation, yielding a decoupled average of 17.22\%. Qwen3-VL-8B-Instruct~\cite{bai2025qwen3vltechnicalreport} and Qwen-VL-Max~\cite{alibabacloud2026qwenmodels} obtain similarly low decoupled averages of 13.70\% and 19.84\%, respectively. The error distribution reveals a consistent bias: models classify pure rotation as translation plus rotation, with misclassification rates of 71.26\%, 99.70\%, and 100.00\% for the three models, respectively. They also confuse pure translation with either translation plus rotation or pure rotation. Findings demonstrate current VLMs conflate viewpoint changes with physical displacement, confirming a severe Visual-Kinematic Misalignment.

\section{Dataset}

Existing spatial datasets entangle visual and motion streams, failing to isolate rotation, translation, and speed. Consequently, models exploit pixel biases rather than learning camera dynamics. We propose the Dyn-3D benchmark. By generating counterfactual videos with identical paths but varying dynamics, Dyn-3D isolates visual variables to assess causal spatial understanding.

\subsection{Data Acquisition and Generation}
\paragraph{Scene Acquisition and Reconstruction} We filter 451 indoor scenes from ScanNet++~\cite{yeshwanthliu2023scannetpp}. To meet the pinhole camera assumption of 3D Gaussian Splatting (3DGS)~\cite{kerbl_3d_2023}, we use OpenCV to undistort fisheye images, standardize resolution to $1752 \times 1168$, and convert COLMAP~\cite{schonberger2016structure} poses into binary format. Each scene retains an average of 261 images. We train each scene for 30,000 iterations using the \texttt{splatfacto} implementation in nerfstudio~\cite{tancik2023nerfstudio} on dual RTX 4090 GPUs. After reconstruction and rendering quality inspection, we retain 447 scenes. A subsequent dataset-level validation yields 443 eligible scenes, comprising 263 training scenes, 167 held-out evaluation scenes, and 13 reserved scenes.

\paragraph{Multi-Trajectory Video Rendering} We generate 30 FPS videos with 5 motion characteristics for each scene. First, we generate a smooth path via K-means clustering of camera poses and cubic spline interpolation. We construct a control group by varying the frame sampling rate along this path: fast (150 frames, 5 seconds), smooth (450 frames, 15 seconds), and slow (900 frames, 30 seconds). Additionally, we synthesize two decoupled trajectories: a rotation trajectory with zero physical displacement but visual changes (360-degree in-place rotation 0.1 meters above the scene center, 120 frames, 4 seconds), and a translation trajectory with a stable viewpoint but physical displacement (directional translation covering 35\% of the spatial depth toward the scene center, 120 frames, 4 seconds).

\paragraph{3D Metadata Extraction} For each video, we extract metadata of objects from the ScanNet++ meshes. To ensure geometric consistency against the scaling and translation operations applied during 3DGS training, we apply inverse transformations using the parameters to restore the 3DGS space to the world coordinate system. For frame-level visibility, we project the 8 vertices of each bounding box into the viewing frustum using camera extrinsics; an object is considered visible if at least one vertex is included. After removing 23 non-interactive background categories (e.g., walls), the system outputs a metadata file for each video. This file logs the scene ID, trajectory type, frame count, resolution, intrinsics, camera motion statistics (total path length, displacement magnitude, accumulated rotation angle, start/end positions), object lists (3D coordinates, bounding boxes, visible frames, distance and direction across keyframes), inter-object distance matrices, and frame-wise camera poses.

\subsection{Annotation Pipeline and Quality Control}

\paragraph{Question Generation} All questions in Dyn-3D are generated deterministically from 3D metadata. The benchmark contains 19 multiple-choice question types (B1 to B19), each with four options, across three dimensions. Kinematic perception (B1--B6 and B19) assesses displacement magnitude, camera-motion direction, rotation, speed, total path length, and the rotation/translation diagnostic traps. Spatial understanding (B7--B16) evaluates object directions, distance trends, nearest-object reasoning, coordinate re-anchoring, counting, size, cross-frame causal direction, and depth-subtraction visibility. Implicit trajectory reasoning (B17--B18) assesses motion type and trajectory shape. Distractors are generated using type-specific strategies, including range partitioning for continuous quantities and sampled alternative directions for directional questions. Classifications are provided in the supplementary material.

\paragraph{Quality Control} We apply three quality-control stages to the evaluation set. First, automated cross-validation independently recomputes each answer from the 3D metadata; all 16,063 aligned questions are marked as geometry-recomputed and validation-passed. Second, physical-validity filtering excludes samples with invalid geometry, ambiguous object references, or out-of-range motion and spatial measurements. Third, representative samples are manually reviewed for question clarity, option validity, and agreement between the rendered evidence and the ground truth.

\begin{table*}[t]
    \centering
    \small
    \setlength{\tabcolsep}{4pt}
    \renewcommand{\arraystretch}{1.15}
    \caption{
    Comprehensive comparison between the Dyn-3D Benchmark and mainstream
    spatial/video understanding benchmarks, including
    ScanQA~\cite{azuma2022scanqa},
    SpatialVLM~\cite{chen2024spatialvlm},
    MVBench~\cite{li2024mvbench},
    EgoSchema~\cite{DBLP:conf/nips/MangalamAM23},
    OpenEQA~\cite{majumdar2024openeqa}, and
    VSI-Bench~\cite{yang2025thinking}.
    ``Counterfactual Render.'' denotes the synthesis of identical spatial
    paths with distinct dynamic characteristics to mitigate shortcut learning.
    }
    \label{tab:dataset_comparison}

    \resizebox{\textwidth}{!}{%
    \begin{tabular}{l c c c c c c c}
        \toprule
        \textbf{Benchmark}
        & \textbf{Modality}
        & \textbf{Videos}
        & \textbf{QAs}
        & \textbf{3D Spatial}
        & \textbf{Ego-Motion}
        & \textbf{Counterfactual Rendering}
        & \textbf{Diagnostic Traps} \\
        \midrule

        ScanQA
        & Static 3D
        & --
        & 41K
        & $\times$
        & $\times$
        & $\times$
        & $\times$ \\

        SpatialVLM
        & Static 2D
        & --
        & 2B
        & \checkmark
        & $\times$
        & $\times$
        & $\times$ \\

        MVBench
        & Video
        & 4K
        & 4K
        & $\times$
        & $\times$
        & $\times$
        & $\times$ \\

        EgoSchema
        & Video
        & 5K
        & 5K
        & $\times$
        & $\times$
        & $\times$
        & $\times$ \\

        OpenEQA
        & Video
        & $\sim$200
        & 1.6K
        & \checkmark
        & $\times$
        & $\times$
        & $\times$ \\

        VSI-Bench
        & Video
        & 288
        & 5K
        & \checkmark
        & Partial
        & $\times$
        & $\times$ \\

        \textbf{Dyn-3D (ours)}
        & \textbf{Video}
        & \textbf{835}
        & \textbf{16K}
        & \textbf{\checkmark}
        & \textbf{\checkmark}
        & \textbf{\checkmark}
        & \textbf{\checkmark} \\

        \bottomrule
    \end{tabular}%
    }
\end{table*}

%The training pool comprises 263 indoor scenes and is strictly disjoint from the 167 held-out scenes in the Dyn-3D Benchmark. The 9,600 SFT samples are drawn from 240 scenes within this training pool; the RL data covers all 263 training scenes. Thus, neither SFT nor RL shares a scene with the benchmark.

%The current experiments use 9,600 SFT samples, 24,000 RL training samples from 263 training scenes. The held-out
%, with five trajectory types per scene: fast, smooth, slow, rotation, and translation. All rendered videos use a resolution of $1752 \times 1168$ at 30 FPS. Fast, smooth, and slow trajectories contain 150, 450, and 900 frames, respectively; rotation and translation trajectories each contain 120 frames.
\subsection{Dataset Statistics and Splits}
The Dyn-3D benchmark consists of 16,063 four-option questions sourced from 167 scenes and 835 rendered videos. It evaluates three dimensions: kinematic perception (3,023 samples, 18.8\%), spatial understanding (11,626 samples, 72.4\%), and implicit trajectory reasoning (1,414 samples, 8.8\%). The dataset is balanced across five trajectory types, with 2,911 to 3,477 questions per trajectory (a maximum-to-minimum ratio of 1.19). Additionally, it includes 168 diagnostic trap questions (147 for rotation and 21 for translation). 
All benchmark answers are independently recomputed from the structured 3D metadata before evaluation.

\subsection{Comparison with Existing Benchmarks}

As shown in Table~\ref{tab:dataset_comparison}, existing benchmarks overlook ego-motion as an explicit variable in spatial cognition. ScanQA and SpatialVLM primarily focus on static spatial understanding, while OpenEQA and VSI-Bench introduce embodied or video-based spatial QA~\cite{azuma2022scanqa,chen2024spatialvlm,majumdar2024openeqa,yang2025thinking}. However, these benchmarks do not explicitly decouple viewpoint rotation, translation, and speed under controlled 3D trajectories.
To address this, we introduce the Dyn-3D benchmark to evaluate VLM ego-motion perception in 3D spaces. Using 3D Gaussian Splatting~\cite{kerbl_3d_2023}, we synthesize counterfactual videos with identical spatial paths but distinct dynamics, including pure rotation and translation traps. This prevents models from exploiting superficial pixel variations, rigorously assessing their kinematic perception and vulnerability to Kinematic Collapse.

\section{Method}

\begin{figure*}[t]
	\centering
	\includegraphics[width=\textwidth]{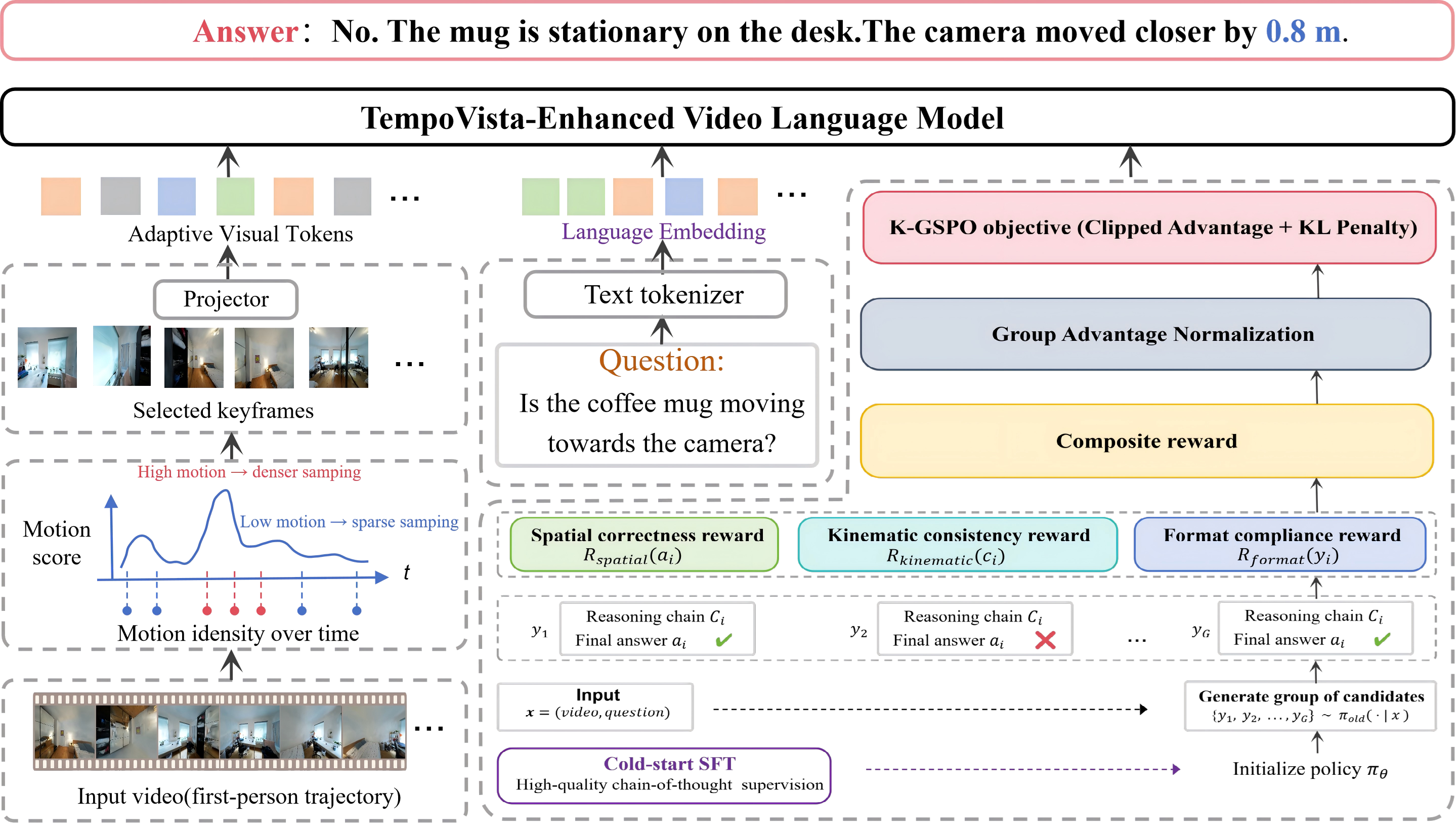}
	\caption{\textbf{Overview of the TempoVista framework.} (Left) The kinematic-adaptive strategy samples keyframes in a motion-aware metric space, yielding denser temporal allocation in segments with larger kinematic variation while preserving critical geometric cues. (Right) The Kinematic-GSPO algorithm optimizes the VLM policy by generating candidate reasoning chains and evaluating them through a composite reward (spatial answer correctness, kinematic consistency, and format compliance), explicitly grounding visual cues in physical 3D space.}
	\label{fig:method}
\end{figure*}

Motivated by the findings from the Dyn-3D benchmark, we propose the TempoVista framework (illustrated in Figure~\ref{fig:method}). Rather than forcing Vision-Language Models (VLMs) to simply memorize 3D data, TempoVista introduces implicit kinematic perception through explicit policy optimization. The framework consists of two key components: a kinematic-adaptive frame selection strategy and the Kinematic-GSPO training algorithm.

\subsection{Kinematic-Adaptive Frame Selection}
\label{sec:frame_selection}
Before feeding long videos into a VLM, keyframe selection determines the upper bound of available geometric information. Traditional uniform sampling assumes information is linearly distributed over time, which may underrepresent intervals where camera motion changes rapidly, such as sharp turns or fast translations. Therefore, we investigate a selection paradigm that allocates the frame budget according to kinematic variation rather than timestamp spacing.

\subsubsection{Oracle Selection via SE(3) FPS}
When reliable camera extrinsics are available (e.g., from odometry or our 3D Gaussian Splatting pipeline), we formulate keyframe selection as a maximum diversity subset selection problem on the Riemannian manifold $SE(3)$.

\noindent\textbf{Distance Metric.} Let $T_i = (R_i,\,\mathbf{t}_i) \in SE(3)$ denote the camera-to-world pose of the $i$-th frame. We define a normalized geodesic-inspired distance:
\begin{equation}
\begin{split}
	d(T_i,\, T_j) &= \alpha \min\left(
	\frac{\bigl\|\mathbf{t}_i - \mathbf{t}_j\bigr\|_2}{D_{\max}}, 1
	\right) \\
	&\quad + (1-\alpha)
	\frac{1}{\pi}
	\arccos\!\Bigl(\tfrac{\operatorname{tr}(R_i^{\top}R_j)-1}{2}\Bigr),
\end{split}
	\label{eq:se3_dist}
\end{equation}
where $D_{\max}$ is a video-level translation normalization constant, the rotation term is the matrix geodesic angle normalized to $[0,1]$, and $\alpha \in (0,1)$ balances translation and rotation.

\noindent\textbf{Greedy FPS.} Given $N$ candidate frames $\{T_i\}_{i=1}^{N}$ and a budget of $k$ keyframes, we initialize the selected set with the first and last frames,
$\mathcal{F}_2 = \{1,N\}$, to preserve temporal boundary conditions, and iteratively append:
\begin{equation}
\begin{split}
	f_{m+1} &= \operatorname*{arg\,max}_{j \notin \mathcal{F}_m} \;\min_{f \in \mathcal{F}_m} d(T_j,\, T_f), \\
	&\qquad m = 2, \dots, k-1.
\end{split}
	\label{eq:fps}
\end{equation}
This algorithm selects the frame farthest from all currently selected frames. The resulting subset forms a geometric skeleton of the trajectory, maximizing the minimum pairwise physical distance in $SE(3)$. Since rapidly changing intervals span larger distances in the kinematic metric space, the selected frames are naturally allocated more densely in such regions, while near-static intervals receive fewer redundant samples. Unlike threshold-based kinematic methods that may over-cluster locally, FPS maintains global trajectory coverage and helps preserve multi-object visibility.

\subsection{Kinematic-GSPO: Physics-Guided Group Sequence Policy Optimization}

We initialize the policy with answer-supervised SFT. During reinforcement learning, the model is encouraged to produce a structured reasoning trace containing five camera-motion quantities---total path length, displacement magnitude, displacement vector, accumulated rotation angle, and average speed---together with the final answer. While answer-level supervision is necessary for spatial question answering, it cannot distinguish a correct answer supported by a kinematically consistent motion interpretation from one obtained through incidental correlations. We therefore introduce \textbf{Kinematic-GSPO}, which injects scene-derived kinematic annotations as process-level feedback during policy optimization.

Given a multi-frame video and a question $x=(\mathcal{V},q)$, we sample a group of $G$ candidate responses $\{y_i\}_{i=1}^{G}$ from the policy $\pi_\theta$. Each response $y_i=(c_i,a_i)$ consists of a structured kinematic reasoning trace $c_i$ and a final answer $a_i$. We define the total reward as
\begin{equation}
\begin{split}
R(y_i;x) &=
R_{\text{ans}}(a_i,a^*)
+ \gamma R_{\text{fmt}}(y_i) \\
&\quad + \lambda R_{\text{kin}}(c_i,\mathbf{k}^*),
\end{split}
\label{eq:kinematic_reward}
\end{equation}
where $R_{\text{ans}}$ rewards answer correctness, $R_{\text{fmt}}$ is a lightweight structural reward that encourages the required XML-style response schema, and $\mathbf{k}^*$ denotes the scene-derived kinematic annotations.

Crucially, $R_{\text{kin}}$ evaluates whether the kinematic quantities declared in $c_i$ are consistent with $\mathbf{k}^*$. For scalar quantities, we use range-consistent matching based on task-relevant motion intervals. For the displacement vector, we jointly evaluate magnitude consistency and directional consistency. The detailed definitions of the five kinematic rewards are provided in the Appendix. This design provides direct process-level supervision on the generated motion trace, encouraging the model to produce spatial answers that are consistent with the underlying camera motion.

We optimize the policy using group-relative advantages. For a sampled group, the advantage of response $y_i$ is normalized by the group reward statistics:
\begin{equation}
A_i =
\frac{R(y_i;x)-\mu_R}
{\sigma_R+\epsilon},
\end{equation}
where $\mu_R$ and $\sigma_R$ are the mean and standard deviation of rewards within the group, respectively. We then optimize a clipped GSPO surrogate objective:
\begin{equation}
\begin{split}
\mathcal{L}_{\text{Kinematic-GSPO}}
= -\frac{1}{G}\sum_{i=1}^{G}
\min \Big[
&A_i \rho_i^{\text{seq}}, \\
&A_i \,
\operatorname{clip}\big(
\rho_i^{\text{seq}},
1-\epsilon_{\ell},
1+\epsilon_{h}
\big)
\Big] \\
&+ \beta D_{\mathrm{KL}}^{\text{seq}}
\left(
\pi_\theta \Vert \pi_{\mathrm{ref}}
\right),
\end{split}
\label{eq:kinematic_gspo}
\end{equation}
where $\rho_i^{\text{seq}}$ is the sequence-level importance ratio, $\epsilon_{\ell}$ and $\epsilon_{h}$ are the lower and upper clipping thresholds, and $\beta$ controls the KL regularization toward the reference policy $\pi_{\mathrm{ref}}$. By combining answer-level supervision with kinematic process rewards, Kinematic-GSPO favors responses whose explicit motion interpretation is consistent with scene-derived physical annotations.
\begin{table}[t]
	\centering
	\small
	\caption{Quantitative evaluation on the Dyn-3D Benchmark using eight frames selected per video. }%Results are reported as accuracy (\%). Best results among open-source models are highlighted in \textbf{bold}.
	\label{tab:main_results}
	\resizebox{\linewidth}{!}{
	\begin{tabular}{l c c c c}
		\toprule
		\textbf{Model}
		& \textbf{Kinematic}
		& \textbf{Spatial}
		& \textbf{Trajectory}
		& \textbf{Overall} \\
		\midrule
		
		\multicolumn{5}{l}{\textit{Proprietary Models}} \\
		GPT-4o
        & 45.3
        & 36.1
        & 40.2
        & 38.2 \\
		
		Qwen-VL-Max
        & 53.2
        & 46.2
        & 71.9
        & 49.8 \\
		
		\midrule
		
		\multicolumn{5}{l}{\textit{Open-Source Models}} \\
		LLaVA-Video
        & 40.7
        & 28.4
        & 61.8
        & 33.7 \\
		
		LLaVA-NeXT-Video
        & 19.1
        & 30.5
        & 34.3
        & 28.7 \\
		
		LLaVA-OneVision
        & 40.3
        & 28.1
        & 56.6
        & 32.9 \\
		
		Qwen2.5-VL-7B-Instruct
        & 30.7
        & 37.4
        & 44.3
        & 36.7 \\
		
		\midrule
		
		InternVL-3.5-8B
		& 52.0
		& 47.8
		& 70.4
		& 50.6 \\
		
		\rowcolor{gray!15}
		InternVL-3.5-8B + TempoVista
		& \textbf{64.5}\,\textcolor{green!60!black}{\scriptsize (+12.5\,$\uparrow$)}
		& 52.7\,\textcolor{green!60!black}{\scriptsize (+4.9\,$\uparrow$)}
		& 77.7\,\textcolor{green!60!black}{\scriptsize (+7.3\,$\uparrow$)}
		& 57.1\,\textcolor{green!60!black}{\scriptsize (+6.5\,$\uparrow$)} \\
		
		Qwen3-VL-8B-Instruct
		& 51.0
		& 47.6
		& 71.4
		& 50.3 \\
		
		\rowcolor{gray!15}
		Qwen3-VL-8B-Instruct + TempoVista
		& 62.6\,\textcolor{green!60!black}{\scriptsize (+11.6\,$\uparrow$)}
		& \textbf{58.3}\,\textcolor{green!60!black}{\scriptsize (+10.7\,$\uparrow$)}
		& \textbf{81.8}\,\textcolor{green!60!black}{\scriptsize (+10.4\,$\uparrow$)}
		& \textbf{61.2}\,\textcolor{green!60!black}{\scriptsize (+10.9\,$\uparrow$)} \\
		
		\bottomrule
	\end{tabular}
	}
\end{table}

\section{Experiments}
\subsection{Dyn-3D-Instruct Dataset}
Dyn-3D-Instruct contains the Supervised Fine-Tuning (SFT) and Reinforcement Learning (RL) data used for training. 
\textbf{SFT Dataset}: The SFT set contains 9,600 multi-frame multiple-choice QA samples from 1,128 rendered videos. Each sample comprises eight video frames, a question with candidate options, and its ground-truth answer. The data comprises six multiple-choice task types: nearest-object identification based on relative distance; four-way, three-way, and binary object-relative direction reasoning; appearance-order tracking; and route-turn prediction. We remove samples with invalid object references, ambiguous geometry, repeated questions, or inconsistent answer-option mappings.
\textbf{RL Dataset}: The RL set contains 24,000 training samples from 1,315 unique rendered videos, with eight frames per sample. The resulting training set covers 24 task types, including camera motion, displacement, rotation, speed, object direction, distance, counting, size, appearance order, coordinate transformation, and route reasoning. Each item is paired with the correct option and fixed five-field kinematic metadata used by Kinematic-GSPO: total path length $d_{\mathrm{path}}$, displacement magnitude $d_{\mathrm{disp}}$, displacement vector $\Delta\mathbf{p}$, accumulated rotation angle $\theta$, and average speed $v$. We verify media paths, answer-option consistency, object-reference validity, and geometric plausibility before training.

\subsection{Experimental Setup}

\noindent\textbf{Evaluation Datasets.}
We evaluate the core spatiotemporal reasoning capabilities of the models on the Dyn-3D benchmark. Its test set contains scenes and counterfactual trajectories that are strictly unseen during training, allowing us to assess generalization beyond memorized visual patterns. In addition, we evaluate on selected multiple-choice spatial understanding tasks from VSI-Bench~\cite{yang2025thinking}, including relative distance estimation, relative direction reasoning, appearance-order tracking, and route planning.

\noindent\textbf{Baselines.}
We compare TempoVista with both proprietary and open-source Vision-Language Models (VLMs). The proprietary baselines include GPT-4o~\cite{openai_gpt4o_2024}, Gemini-1.5-Flash, Gemini-1.5-Pro~\cite{google_gemini_1.5_2024}, and Qwen-VL-Max~\cite{alibabacloud2026qwenmodels}. The open-source baselines include representative video understanding models from the LLaVA family~\cite{li2024llava,li2025llavanextinterleave,damonlpsg2024videollama2}, Qwen2.5-VL~\cite{bai2025qwen25vltechnicalreport}, InternVL-3.5~\cite{chen2024internvlscalingvisionfoundation}, and Qwen3-VL~\cite{bai2025qwen3vltechnicalreport}. We instantiate TempoVista using two base models, InternVL-3.5-8B and Qwen3-VL-8B-Instruct, to examine whether the proposed framework generalizes across different VLM architectures.

\subsection{Kinematic Perception Evaluation} Table~\ref{tab:main_results} reports the Dyn-3D results using eight selected frames per video. Existing open-source video-language models, particularly the LLaVA series, achieve only 28.7\%--33.7\% overall accuracy, demonstrating the difficulty of fine-grained 3D motion reasoning. TempoVista consistently improves both evaluated base models. For InternVL-3.5-8B, it raises the Kinematic, Spatial, Trajectory, and Overall accuracies from 52.0\%, 47.8\%, 70.4\%, and 50.6\% to 64.5\%, 52.7\%, 77.7\%, and 57.1\%, respectively. This corresponds to gains of 12.5, 4.9, 7.3, and 6.5 percentage points, with the best open-source Kinematic result. When applied to Qwen3-VL-8B-Instruct, TempoVista improves the corresponding results from 51.0\%, 47.6\%, 71.4\%, and 50.3\% to 62.6\%, 58.3\%, 81.8\%, and 61.2\%, yielding gains of 11.6, 10.7, 10.4, and 10.9 points. This variant achieves the best open-source Spatial, Trajectory, and Overall results and surpasses Qwen-VL-Max by 11.4 points overall. The consistent gains across both model families indicate that TempoVista generalizes across VLM architectures, while their complementary strengths suggest that InternVL benefits more in direct kinematic perception and Qwen3-VL benefits more in broader spatial and trajectory reasoning.

\begin{table}[t]
	\centering
	\caption{Evaluation results on the selected VSI-Bench MCQ spatial understanding tasks using 32 frames per video. }%Best results are highlighted in \textbf{bold}.
	\label{tab:vsi_results}
	\resizebox{\linewidth}{!}{
	\begin{tabular}{l c c c c c}
		\toprule
		\textbf{Model} & \textbf{Overall} & \textbf{Rel. Distance} & \textbf{Rel. Direction} & \textbf{App. Order} & \textbf{Route Plan} \\
		\midrule
		\multicolumn{6}{l}{\textit{Proprietary Models}} \\
		GPT-4o & 36.1 & 37.0 & 41.3 & 28.5 & 31.5 \\
		Gemini-1.5-Flash & 38.5 & 37.7 & 41.0 & 37.8 & 31.5 \\
		Gemini-1.5-Pro & 44.0 & 51.3 & 46.3 & 34.6 & \textbf{36.0} \\
		\midrule
		\multicolumn{6}{l}{\textit{Open-Source Models}} \\
		Qwen2.5-VL-7B-Instruct & 34.5 & 38.0 & 37.4 & 28.0 & 28.4 \\
		LLaVA-OneVision-7B & 34.1 & 42.5 & 35.2 & 24.4 & 29.4 \\
		LLaVA-NeXT-Video-7B & 39.1 & 43.5 & 42.4 & 30.6 & 34.0 \\
		\midrule
		InternVL-3.5-8B
		& 50.1
		& 52.4
		& 48.6
		& 55.1
		& 32.9 \\

		\rowcolor{gray!15}
		InternVL-3.5-8B + TempoVista
		& 50.7\,\textcolor{green!60!black}{\scriptsize (+0.6\,$\uparrow$)}
		& 54.4\,\textcolor{green!60!black}{\scriptsize (+2.0\,$\uparrow$)}
		& 48.1\,\textcolor{red}{\scriptsize (-0.5\,$\downarrow$)}
		& 55.7\,\textcolor{green!60!black}{\scriptsize (+0.6\,$\uparrow$)}
		& 34.5\,\textcolor{green!60!black}{\scriptsize (+1.6\,$\uparrow$)} \\

		Qwen3-VL-8B-Instruct
		& 51.3
		& 53.5
		& 47.9
		& 60.2
		& 32.0 \\

		\rowcolor{gray!15}
		Qwen3-VL-8B-Instruct + TempoVista
		& \textbf{55.6}\,\textcolor{green!60!black}{\scriptsize (+4.3\,$\uparrow$)}
		& \textbf{56.2}\,\textcolor{green!60!black}{\scriptsize (+2.7\,$\uparrow$)}
		& \textbf{51.7}\,\textcolor{green!60!black}{\scriptsize (+3.8\,$\uparrow$)}
		& \textbf{67.8}\,\textcolor{green!60!black}{\scriptsize (+7.6\,$\uparrow$)}
		& 34.0\,\textcolor{green!60!black}{\scriptsize (+2.0\,$\uparrow$)} \\
		
		\bottomrule
	\end{tabular}
	}
\end{table}

\subsection{Spatial Understanding Evaluation} Table~\ref{tab:vsi_results} evaluates transfer to selected VSI-Bench spatial reasoning tasks. TempoVista improves InternVL-3.5-8B from 50.1\% to 50.7\% overall, with gains in relative distance, appearance order, and route planning, although relative-direction accuracy decreases slightly from 48.6\% to 48.1\%. For Qwen3-VL-8B-Instruct, it improves overall accuracy from 51.3\% to 55.6\%, including gains of 2.7, 3.8, 7.6, and 2.0 points in relative distance, relative direction, appearance order, and route planning, respectively. Qwen3-VL-8B-Instruct + TempoVista therefore obtains the best open-source results on all evaluated categories except route planning, where InternVL-3.5-8B + TempoVista reaches 34.5\%. These results show that kinematic-aware optimization generally transfers to external spatial reasoning tasks, although the magnitude and consistency of the gains remain dependent on the base model and task category.

\subsection{Ablation Study}

\noindent\textbf{Training Strategy.}
Table~\ref{tab:ablation_training} compares different training strategies on two base models. For InternVL-3.5-8B, SFT improves accuracy from 50.6\% to 51.6\%, while Base GSPO further increases it to 55.5\%. Kinematic-GSPO achieves 57.1\%, outperforming Base GSPO by 1.6 percentage points and the original base model by 6.5 points. For Qwen3-VL-8B-Instruct, SFT, Base GSPO, and Kinematic-GSPO achieve 52.6\%, 60.1\%, and 61.2\%, respectively, compared with 50.3\% for the base model. These results show that policy optimization provides the majority of the improvement, while the proposed kinematic reward consistently contributes an additional gain of 1.1--1.6 points across both model families.

\noindent\textbf{Sampling Strategy.}
Table~\ref{tab:ablation_sampling} compares the oracle SE(3)-based sampling strategy adopted by TempoVista with uniform sampling and a flow-proxy ablation under non-linear trajectories. For InternVL-3.5-8B + TempoVista, uniform, flow-proxy, and oracle sampling achieve 55.6\%, 55.1\%, and 57.1\%, respectively. The corresponding results for Qwen3-VL-8B-Instruct + TempoVista are 59.7\%, 59.0\%, and 61.2\%. The adopted oracle strategy consistently performs best, exceeding uniform sampling by 1.5 points for both models, whereas the flow-proxy variant underperforms uniform sampling by 0.5 and 0.7 points, respectively. These results confirm that frame selection based on accurate SE(3) camera trajectories is more effective than either uniform sampling or the visual-motion proxy.

\begin{table}[t]
\centering
\begin{minipage}[t]{0.48\textwidth}
	\centering
	\small
	\caption{Ablation on training strategies with InternVL-3.5-8B and Qwen3-VL-8B-Instruct as base models.}
	\label{tab:ablation_training}
	\begin{tabularx}{\linewidth}{l X c}
		\toprule
		\textbf{ID} & \textbf{Training Strategy} & \textbf{Accuracy (\%)} \\
		\midrule
		
		(a) & {\textit{Base model:} \textbf{InternVL-3.5-8B}}    & 50.6 \\
		(b) & SFT Only      & 51.6 \\
		(c) & Base GSPO     & 55.5 \\
		(d) & Kinematic-GSPO (Ours) & \textbf{57.1} \\
		
		\midrule
		
		(a) & {\textit{Base model:} \textbf{Qwen3-VL-8B-Instruct}}    & 50.3 \\
		(b) & SFT Only      & 52.6 \\
		(c) & Base GSPO     & 60.1 \\
		(d) & Kinematic-GSPO (Ours) & \textbf{61.2} \\
		
		\bottomrule
	\end{tabularx}
\end{minipage}
\hfill
\begin{minipage}[t]{0.48\textwidth}
	\centering
	\small
	\caption{Ablation on frame sampling strategies for InternVL-3.5-8B and Qwen3-VL-8B-Instruct under non-linear trajectories.}
	\label{tab:ablation_sampling}
	\begin{tabularx}{\linewidth}{l X c}
		\toprule
		\textbf{ID} & \textbf{Sampling Strategy} & \textbf{Accuracy (\%)} \\
		\midrule
		
		\multicolumn{3}{l}{\textbf{InternVL-3.5-8B + TempoVista}} \\
		(a) & Uniform & 55.6 \\
		(b) & Flow Proxy & 55.1 \\
		(c) & Oracle & \textbf{57.1} \\
		
		\midrule
		
		\multicolumn{3}{l}{\textbf{Qwen3-VL-8B-Instruct + TempoVista}} \\
		(a) & Uniform & 59.7 \\
		(b) & Flow Proxy & 59.0 \\
		(c) & Oracle & \textbf{61.2} \\
		
		\bottomrule
	\end{tabularx}
\end{minipage}
\end{table}

\section{Conclusion}

We identify kinematic collapse in vision language models, a spatial cognition failure under nonsmooth trajectories caused by reliance on 2D visual shortcuts. To evaluate this, we introduce the Dyn-3D benchmark, which uses counterfactual rendering to decouple visual changes from physical motion. We further propose the TempoVista framework, featuring a kinematic-adaptive frame selection and the Kinematic-GSPO algorithm. By embedding metric motion constraints into policy optimization, Kinematic-GSPO grounds visual cues in 3D space. Experiments show TempoVista outperforms leading open and proprietary models on Dyn-3D and generalizes to diverse spatial tasks in VSI-Bench. These findings prove explicit kinematic perception is essential for robust 3D spatial understanding.

%\section*{Limitations}
%Despite the improvements demonstrated by TempoVista, our current framework has several limitations. First, the Dyn-3D benchmark is built upon static indoor scenes reconstructed via 3D Gaussian Splatting~\cite{kerbl_3d_2023}. It lacks dynamic foreground objects, which prevents the evaluation of independent object motion. Second, the optical flow proxy used for frame selection may become unreliable under severe motion blur or in completely textureless regions. Finally, scaling the training algorithm to unconstrained web videos remains challenging. The current policy optimization relies on accurate spatial metadata and camera poses, which are difficult to extract from standard internet videos. Future work will explore unsupervised physical constraints to extend kinematic perception to highly dynamic real world environments.

% References and End of Paper

\clearpage
\bibliographystyle{plainnat}
\bibliography{main}

\clearpage

\clearpage
\thispagestyle{plain}

\vspace*{0mm}
\noindent{\titlefont\sffamily \seedblue{Authors}\par}
\vspace{5mm}

\noindent
{\sffamily\bfseries\large Author List}

\vspace{3mm}

\noindent
Jiayu Ding$^{*}$, Zhuodong Liu$^{*}$, Lei Zhang$^{*}$, Manyu Xiong, Hongbo Jin, Haoran Tang, Hongbo Zhang, Changen Zhu, Wenbo Xing$^{\dagger}$

\noindent
{\sffamily\small
	$^{*}$ These authors contributed equally to this work.\\
	$^{\dagger}$ Project leader and corresponding author.
}

\vspace{5mm}

\noindent
{\sffamily\bfseries\large Author Contributions}

\vspace{3mm}

\noindent
{\sffamily\bfseries Jiayu Ding.}
Overall method design, model design and training, and paper writing. Co-first author.
{\sffamily\bfseries Zhuodong Liu.}
Construction and evaluation of the benchmark. Co-first author.
{\sffamily\bfseries Lei Zhang.}
Construction of the training data. Co-first author.
{\sffamily\bfseries Manyu Xiong.}
Participated in the evaluation of the benchmark.
{\sffamily\bfseries Hongbo Jin.}
Participated in the overall method design and discussion.
{\sffamily\bfseries Haoran Tang.}
Participated in the overall method design and discussion.
{\sffamily\bfseries Hongbo Zhang.}
Participated in dataset cleaning, verification, and annotation.
{\sffamily\bfseries Changen Zhu.}
Participated in dataset cleaning, verification, and annotation.
{\sffamily\bfseries Wenbo Xing.}
Overall method design, model design and training, and paper writing. Project leader and corresponding author.

\clearpage
\beginappendix

\section{Dataset Construction Details}
\label{sec:dataset_construction_details}

\paragraph{Scene Filtering.}
Before using a scene for dataset construction, we apply a two-stage filtering process to ensure that both semantic annotations and rendered videos are reliable. In the first stage, we start from 499 raw ScanNet++ scenes and retain 451 scenes with complete \texttt{segments\_anno.json} annotations. These annotations provide the ground-truth source for extracting object semantics, 3D bounding boxes, visibility, and object-level spatial relations. Scenes without complete object annotations are discarded because their metadata cannot support deterministic question generation.

In the second stage, we train 3D Gaussian Splatting models and render multi-trajectory videos for all 451 candidate scenes. Each rendered scene is manually inspected to verify reconstruction and video quality. We remove 4 scenes that satisfy any of the following failure conditions: 3DGS training diverges or produces severe geometric distortion; one or more trajectory videos fail to render or contain broken geometry; the rendered frames fall below the quality required for evaluation, including severe floaters, holes, or strong motion blur. After this filtering, we retain 447 high-fidelity scenes as the final data source. A subsequent dataset-level QA, media, and geometry validation excludes four additional scenes, yielding 443 eligible scenes for the final experimental pool. Among them, 263 scenes are used for training and 167 scenes are held out for evaluation, with no scene overlap. The remaining 13 eligible scenes are held in reserve for potential future use.

\paragraph{Reconstruction and Rendering.}
For each retained scene, we first undistort the original fish-eye images using OpenCV to satisfy the pinhole camera assumption of 3D Gaussian Splatting~\cite{kerbl_3d_2023}. We standardize image resolution to $1752 \times 1168$ and convert COLMAP~\cite{schonberger2016structure} camera poses into the required binary format. Each scene keeps an average of 261 valid images. We then train each scene for 30,000 iterations using the \texttt{splatfacto} implementation in nerfstudio~\cite{tancik2023nerfstudio} on dual RTX 4090 GPUs. The trained 3DGS models are used to render videos under controlled camera trajectories.

\paragraph{Counterfactual Trajectory Design.}
Dyn-3D is designed to decouple visual change from physical camera motion. For each scene, we render five trajectory types at 30 FPS. First, we generate a shared smooth physical path by clustering camera poses and applying cubic spline interpolation. Based on this identical spatial path, we vary only the temporal sampling rate to obtain three counterfactual videos: fast, smooth, and slow. The fast trajectory contains 150 frames and lasts 5 seconds, the smooth trajectory contains 450 frames and lasts 15 seconds, and the slow trajectory contains 900 frames and lasts 30 seconds. This design keeps the spatial path unchanged while changing the motion dynamics.

We further synthesize two diagnostic trajectories. The rotation trajectory has almost zero physical displacement but strong visual changes caused by a 360-degree in-place rotation. The translation trajectory has clear physical displacement while keeping the viewpoint relatively stable. These two trajectories are used to test whether models confuse pixel-level visual variation with true metric motion.

\paragraph{3D Metadata Extraction.}
For every rendered video, we extract structured 3D metadata from the ScanNet++ meshes and object annotations. Since 3DGS training may apply scaling and translation, we use the recorded transformation parameters to map the reconstructed scene back to the original metric coordinate system. We compute camera motion statistics, including total path length, displacement magnitude, accumulated rotation angle, start and end positions, and speed. For object-level metadata, we record semantic labels, 3D bounding boxes, object centers, visible frames, keyframe distances, egocentric directions, inter-object distance matrices, and frame-wise camera poses. Frame-level visibility is estimated by projecting the eight vertices of each 3D bounding box into the camera frustum. An object is treated as visible if at least one projected vertex lies inside the valid view. We remove 23 non-interactive background categories, such as walls and ceilings, to focus the benchmark on objects that support spatial reasoning.

\section{Benchmark Details}
\label{sec:benchmark_details}

Dyn-3D contains 19 multiple-choice question types, denoted as B1 to B19. Each question has four options and is generated deterministically from the extracted 3D metadata. The benchmark covers three major dimensions: kinematic perception, spatial understanding, and implicit trajectory reasoning. The detailed definitions are listed below.

\begin{itemize}
    \item \textbf{B1: Displacement Magnitude Estimation.} Given a video, the model estimates the straight-line displacement between the initial and final camera positions.
    \item \textbf{B2: Camera Motion Direction.} The model predicts the egocentric direction of camera displacement, such as front, rear, left, right, or diagonal directions.
    \item \textbf{B3: Rotation Angle Estimation.} The model estimates the accumulated camera rotation angle across the video.
    \item \textbf{B4: Rotation Trap.} The video contains strong visual changes caused mainly by in-place rotation. The model must avoid mistaking rotation-induced appearance changes for large physical displacement.
    \item \textbf{B5: Translation Trap.} The video contains clear physical translation with relatively stable visual appearance. The model must detect true displacement rather than relying only on pixel-level changes.
    \item \textbf{B6: Camera Speed Estimation.} The model estimates the average camera-speed category from the rendered trajectory.
    \item \textbf{B7: Object Direction at the Initial Frame.} The model predicts the egocentric direction of a queried object relative to the camera at the beginning of the video.
    \item \textbf{B8: Object Direction at the Final Frame.} The model predicts the egocentric direction of a queried object relative to the camera at the end of the video.
    \item \textbf{B9: Object Distance Trend.} The model determines whether a queried object becomes closer, farther, or remains at a similar distance as the camera moves.
    \item \textbf{B10: Nearest Object Reasoning.} The model identifies the nearest visible object from a candidate set under a specified frame or trajectory condition.
    \item \textbf{B11: Coordinate Re-Anchoring.} The coordinate system is redefined using a reference object and an orientation axis. The model must infer the new coordinate frame and transform object positions accordingly.
    \item \textbf{B12: Relative Object Distance.} The model compares two or more objects and determines which one is closer to the camera or to a reference object.
    \item \textbf{B13: Object Counting.} The model counts visible objects that satisfy a spatial or semantic condition.
    \item \textbf{B14: Object Size Reasoning.} The model compares the physical sizes of objects using 3D bounding-box metadata.
    \item \textbf{B15: Cross-Frame Causal Direction.} The model infers a spatial direction by comparing object or camera states across temporal positions.
    \item \textbf{B16: Depth-Subtraction Visibility.} The model reasons about whether an object remains visible after accounting for depth and occlusion relations.
    \item \textbf{B17: Motion Type Recognition.} The model infers whether the trajectory is dominated by rotation, translation, or mixed motion.
    \item \textbf{B18: Trajectory Shape Reasoning.} The model identifies the geometric pattern of the camera path, such as straight, curved, or circular motion.
    \item \textbf{B19: Total Path-Length Estimation.} The model estimates the total camera travel distance accumulated along the trajectory.
\end{itemize}

These question types separate physically grounded motion perception from superficial video recognition. B1--B6 and B19 test metric kinematic perception; B7--B16 evaluate object-level spatial understanding under egocentric motion; and B17--B18 require higher-level inference over trajectory patterns. B4 and B5 are diagnostic traps: B4 tests whether models overestimate displacement under large visual changes, while B5 tests whether models underestimate displacement when visual changes are weak. This design probes the failure mode we call \emph{Kinematic Collapse}, in which models fail to maintain a consistent physical interpretation of camera motion.

\section{Method Details}

This section details the calculation of the kinematic reward $R_{\text{kin}}$ used by the fixed-five-field Kinematic-GSPO configuration. It evaluates five ground-truth quantities: $\mathbf{k}^*=(d_{\text{path}}^*, d_{\text{disp}}^*, \Delta\mathbf{p}^*, \theta^*, v^*)$, corresponding to total path length, displacement magnitude, displacement vector, accumulated rotation angle, and average speed. Unless otherwise stated, the total reward is
\begin{equation}
R = R_{\mathrm{ans}} + \gamma R_{\mathrm{fmt}} + \lambda R_{\mathrm{kin}},
\end{equation}
where $R_{\mathrm{ans}}\in\{0,1\}$ is the option-answer reward, $\gamma=0.1$, and $\lambda=0.1$ for Kinematic-GSPO. Base GSPO uses the same setting with $\lambda=0$. $R_{\mathrm{kin}}$ is the unweighted average of all five field-level rewards, where a missing or unparsable field receives a reward of $-1$.

\noindent\textbf{Scalar Metrics.} Predicted and true scalars are quantized into four bins based on specific thresholds: $\{0.5, 4, 9\}$m for total path length $d_{\text{path}}$, $\{0.5, 1.5, 3\}$m for displacement magnitude $d_{\text{disp}}$, $\{15^\circ, 360^\circ, 600^\circ\}$ for accumulated rotation angle $\theta$, and $\{0.15, 0.4, 0.8\}$m/s for average speed $v$. The scalar rewards assign $+1$ for an exact bin match, $0$ for a one-bin error, and $-1$ for larger deviations or missing outputs.

\noindent\textbf{Vector Metric.} For the displacement vector $\Delta\mathbf{p}$, we evaluate both magnitude and direction. Magnitude is scored using the same bin thresholds as $d_{\text{disp}}$. For direction, we calculate the cosine similarity between the predicted and ground-truth vectors: $\cos \ge 0.866$ yields $+1$, $0.5 \le \cos < 0.866$ yields $0$, and $\cos < 0.5$ yields $-1$. The vector reward averages the magnitude and direction scores. If either vector has a near-zero norm, the direction score is $+1$ only when both predicted and ground-truth vectors have norm below $0.5$m; otherwise it is $-1$.

\noindent\textbf{Formatting Weight.} For the formatting reward $R_{\text{fmt}}$ used in the total reward formulation, we empirically set the scaling factor to $\gamma=0.1$.

\subsection{Implementation Details}
\label{sec:implementation_details}

\noindent\textbf{Frame Selection and Input Budget.} For Dyn-3D, each model receives $k=8$ frames per rendered video selected by the oracle SE(3)-FPS procedure. We set the translation weight to $\alpha=0.7$. For each video, $D_{\max}$ is the $95$th percentile of all pairwise camera-center distances plus $10^{-6}$; the translation term is subsequently clipped at one as in Eq.~\ref{eq:se3_dist}. The SFT and RL samples likewise contain eight frames per example. For VSI-MCQ, all reported results use 32 cached frames per video and the 2,490-question MCQ test set.

\noindent\textbf{SFT and Parameter-Efficient Adaptation.} Both model families are first adapted using answer-only SFT on 9,600 eight-frame multiple-choice examples. We train for one epoch with an effective batch size of 16 and learning rate $2\times10^{-5}$. We use LoRA with rank 32 and scaling factor 64 on the language-model projections $\{q,k,v,o,\mathrm{gate},\mathrm{up},\mathrm{down}\}$, while freezing the vision encoder. For InternVL, we additionally freeze the visual projector (\texttt{mlp1}). The SFT adapters are merged with their respective base models before RL. Both Base GSPO and Kinematic-GSPO are initialized from the same merged answer-only SFT checkpoint; they do not start directly from the unadapted base model. During RL, we attach a new LoRA adapter with the same rank, scaling factor, and language-model target projections; the vision encoder remains frozen, and the InternVL visual projector remains frozen.

\noindent\textbf{GSPO Optimization.} Each rollout batch contains 12 prompts, with $G=4$ responses sampled per prompt, resulting in 48 generated responses per rollout batch. Rollouts use temperature $1.0$, top-$p=0.95$, no top-$k$ truncation, and a maximum response length of 192 tokens. We use two GPUs and an update micro-batch size of one per GPU. We optimize with AdamW using learning rate $10^{-5}$, weight decay $10^{-2}$, a cosine schedule with 150 warmup steps, and one policy-update epoch per rollout batch. We set the maximum number of optimization steps to 3,000. The implementation uses the \texttt{gspo\_token} loss with sequence averaging (\texttt{loss\_avg\_mode=seq}). Its forward importance ratio is sequence-level, as represented in Eq.~\ref{eq:kinematic_gspo}, while token-level gradient correction is used during optimization. We use asymmetric clipping thresholds $\epsilon_{\ell}=3\times10^{-4}$ and $\epsilon_{h}=4\times10^{-4}$, and the KL coefficient is $\beta=10^{-2}$. The RL prompt explicitly requests an XML response with a \texttt{<reasoning>} block containing \texttt{truth}, \texttt{cot}, and \texttt{answer} lines; the \texttt{truth} line lists total path length, displacement magnitude, displacement vector, accumulated rotation angle, and average speed with units, followed by an outer \texttt{<answer>} tag. Thus, the structured trace is an instructed training output rather than an unconstrained emergent format.

\noindent\textbf{Reward and Missing Fields.} $R_{\mathrm{fmt}}$ is a format reward rather than a penalty. Its raw value is $1$ when both the reasoning block and outer \texttt{<answer>} tag are present, $0$ when only the outer answer tag is present, $-0.5$ when an answer is parsed without the outer tag, and $-1$ when no answer is parsed. If the reasoning block is absent, the kinematic reward is $-1$; a missing scalar or displacement vector is likewise assigned $-1$ for the corresponding field. The answer reward is one for a correct option and zero otherwise. We use $\gamma=0.1$ and $\lambda=0.1$ for Kinematic-GSPO, while Base GSPO sets $\lambda=0$ and keeps all remaining settings unchanged.

\noindent\textbf{Evaluation Prompt and Parsing.} For Dyn-3D and VSI-MCQ evaluation, the question and options are followed by the instruction \textit{``Please answer with the option letter.''} Generation is greedy. Answers are parsed in the following order: an \texttt{<answer>} tag, an \texttt{Answer:}/\texttt{Option:} letter, an exact standalone option letter, and finally the last standalone option letter in the response. VSI-MCQ additionally maps a normalized free-form response to a uniquely matching option text when no option letter is found. Reported Dyn-3D parse rates refer to this answer-letter parser and should not be interpreted as strict XML-format compliance rates.

\section{Error Analysis}
\label{sec:error_analysis}

We analyze representative failure cases on Dyn-3D and VSI-Bench to characterize the remaining limitations of TempoVista. All Dyn-3D models included in our comparison achieve a 100\% answer parse rate. Therefore, the failures discussed below arise from incorrect spatial or kinematic reasoning rather than invalid response formatting.

On Dyn-3D, TempoVista substantially improves both base models across all three evaluation dimensions. For Qwen3-VL-8B-Instruct, the accuracy increases from 51.0\% to 62.6\% on kinematic perception, from 47.6\% to 58.3\% on spatial understanding, and from 71.4\% to 81.8\% on trajectory reasoning. For InternVL-3.5-8B, the corresponding results improve from 52.0\% to 64.5\%, from 47.8\% to 52.7\%, and from 70.4\% to 77.7\%, respectively. Despite these consistent gains, systematic errors remain in egocentric alignment, coordinate transformation, and metric motion estimation.

\paragraph{Egocentric Depth-Axis Inversion.}
A recurring failure involves inversion of the egocentric depth axis. In one inspected example, the question asks for the direction of a suitcase relative to the camera in the initial frame. The model predicts \emph{Rear}, whereas the ground-truth answer is \emph{Front}. In another example, it predicts \emph{Rear-Left} for a camping bag whose correct relation is \emph{Front-Left}. Because the left--right component is preserved, the model appears to identify the queried object correctly but reverses the front--rear relation.

For questions referring to a single frame, this pattern is more appropriately interpreted as egocentric depth-axis confusion than as a temporal updating failure. In cross-frame settings, however, the same error may also result from incorrect propagation of the camera-centered coordinate frame over time.

\paragraph{Coordinate Re-Anchoring.}
Coordinate re-anchoring remains challenging. In one Dyn-3D example, the coordinate system is defined using the bed as the origin and the direction from the bed to the sofa as the positive $Y$-axis. The correct direction of the phone charger is \emph{Front-Right}, whereas the model predicts \emph{Rear-Left}. Solving this problem requires the model to first ground the referenced objects, construct the requested coordinate system, and then transform the queried object's position into the newly defined frame. The failure suggests that object grounding and explicit coordinate transformation are not yet reliably composed, even when the relevant objects are visually identifiable.

\paragraph{Fine-Grained Motion Estimation.}
Residual errors are also observed in fine-grained motion estimation. In one camera-motion example, the model predicts \emph{Rear} instead of \emph{Rear-Right}. It therefore recovers the dominant backward component but fails to identify the lateral displacement. In another example, a near-stationary trajectory is classified as having \emph{Moderate} speed. These cases indicate that visible image variation can still be confused with actual camera displacement or physical speed. Although the kinematic supervision substantially improves aggregate performance, it does not completely resolve fine-grained metric estimation under subtle or compound motion.

\paragraph{Transfer to Real-World Spatial Reasoning.}
Similar reasoning failures occur on VSI-Bench. For example, when determining whether a sofa is front-left, front-right, back-left, or back-right from a stove while facing a television, the model predicts \emph{Front-Left} instead of the correct answer, \emph{Front-Right}. In a route-planning example, an incorrect initial turn propagates to the subsequent navigation decisions, leading to an incorrect final route.

For Qwen3-VL-8B-Instruct, TempoVista improves the VSI-Bench overall accuracy from 51.3\% to 55.6\%. Relative-distance accuracy increases from 53.5\% to 56.2\%, relative-direction accuracy from 47.9\% to 51.7\%, appearance-order accuracy from 60.2\% to 67.8\%, and route-planning accuracy from 32.0\% to 34.0\%. The largest gain is obtained on appearance-order reasoning, with an improvement of 7.6 percentage points.

For InternVL-3.5-8B, the overall accuracy increases more modestly from 50.1\% to 50.7\%. Relative-distance accuracy improves from 52.4\% to 54.4\%, appearance-order accuracy from 55.1\% to 55.7\%, and route-planning accuracy from 32.9\% to 34.5\%. In contrast, relative-direction accuracy decreases slightly from 48.6\% to 48.1\%. These results indicate that transfer to general spatial understanding is positive overall but remains dependent on both the underlying base model and the specific spatial reasoning task.

Overall, the qualitative failures and quantitative results show that improved motion grounding can co-occur with stronger spatial reasoning. However, kinematic supervision alone does not solve every spatial operation. In particular, explicit coordinate re-anchoring, egocentric depth disambiguation, and fine-grained metric motion estimation remain important directions for future improvement.

\end{document}